\documentclass[12pt]{article}

\usepackage[english]{babel}

\usepackage[a4paper,top=1in,bottom=1in,left=1in,right=1in]{geometry}
\usepackage[singlespacing]{setspace}

\providecommand{\keywords}[1]
{
  \small	
  \textbf{\textit{Keywords---}} #1
}

\usepackage{amsmath}
\usepackage{graphicx}
\usepackage[colorlinks=true, allcolors=blue]{hyperref}
\title{Making AI `Smart': Bridging AI and Cognitive Science}
\author{Madhav Agarwal, Siddhant Bansal}
\date{International Institute of Information Technology, Hyderabad
}\begin{document}
\maketitle

\begin{abstract}
The last two decades have seen tremendous advances in Artificial Intelligence. The exponential growth in terms of computation capabilities has given us hope of developing humans like robots. The question is: are we there yet? Maybe not. With the integration of cognitive science, the `artificial' characteristic of Artificial Intelligence (AI) might soon be replaced with `smart'. This will help develop more powerful AI systems and simultaneously gives us a better understanding of how the human brain works. We discuss the various possibilities and challenges of bridging these two fields and how they can benefit each other. We argue that the possibility of AI taking over human civilization is low as developing such an advanced system requires a better understanding of the human brain first. 
\end{abstract}


\section{Introduction}

The idea of neural networks dates back to 1943 when neurophysiologist Warren McCulloch and mathematician Walter Pitts wrote a paper on how neurons might work and demonstrated it using electric circuits~\cite{mcculloch1943logical}. 
In the initial phase, the idea of neural networks and perceptron gain popularity (Rosenblatt~\cite{rosenblatt1958perceptron}, Widrow \& Hoff~\cite{widrow1960adaptive}). 
Soon this hype hit a major roadblock with the publication of a book, “Perceptrons” by Marvin Minsky and Seymour Papert~\cite{marvin1969perceptrons}. 
The book concluded that perceptron could not be translated effectively into multi-layered neural networks and hence begin the `the AI winter'. 
Fast-forwarding it to 1985, the work of Rumelhart, Hinton, and Williams~\cite{rumelhart1985learning,rumelhart1986learning} gave new life to the neural networks. 
Furthermore, improvement in the computation capabilities enabled the researchers to train large neural networks and beat the current state-of-the-art rule-based methods.
The first feat was the success of AlexNet~\cite{krizhevsky2012imagenet}, which significantly improved on image classification task.
With the beginning of the `deep learning' era, neural networks have gained popularity in multiple domains, including object and face recognition (He,Zhang, Ren, \& Sun~\cite{he2016deep}, and Lu \& Tang~\cite{lu2015surpassing}), speech and natural language processing (Amodei, Dario, et al.~\cite{amodei2016deep} and Ferrucci \& David A.~\cite{ferrucci2012introduction}). 
The networks are matching or even beating humans in complex cognitive tasks such as playing go or video games (Mnih et al.~\cite{mnih2015human}, and  Silver et al.~\cite{silver2016mastering}). The success makes us hopeful that it is possible to make computers smart enough to mimic human cognitive abilities with extensive data, better hardware, and more generalizable models.
In fact, some of the models have surpassed the classical theories in linguistics and physiology to achieve success. 
However, such cases are rare, and they lack a complete fundamental understanding of how the model is learning.

Another interesting thing to note here is that, majority of models' algorithms and architectures are loosely inspired by human cognition.
Cognitive Science and AI have shared a long history, each benefiting the other. Empirical data from cognitive psychology has played an essential role in identifying the failure modes of current AI systems (Nematzadeh et al.~\cite{nematzadeh2018evaluating}, Hamrick et al.~\cite{hamrick2019analogues}). 
Similarly, methods and data from AI have helped in better understanding of human learning by building models that mimic human cognition (Tenenbaum et al.~\cite{tenenbaum2006theory}, Lieder \& Griffiths~\cite{lieder2017strategy}). 

In this paper, we explore the various developments in the field of AI and cognitive science. We argue how a better understanding of the brain can help improve the AI models and how AI needs to move from mere `pattern recognition' to actually understanding the world. Additionally, we discuss some of the key components which will help in building more human-like machines. Furthermore, we look into the various challenges that impede the researchers in these communities from working more closely.  Also, we discuss the possibility of AI taking over human intelligence.
Finally, we conclude by looking at various common themes on which AI and cognitive science collaboration can do wonders.

\section{AI is everywhere}

AI has become an integral part of human life. From using Google Maps for road navigation to using social networking sites, we are taking advantage of the developments in the field of AI.
We have developed `smart' personal assistants like Alexa and Siri, that help with various aspects of our day-to-day life.
Other recent gifts of AI are autonomous cars, autonomous drones, assistive robots, etc. All this has become possible due to the introduction of the Artificial Neural Networks (ANNs).
On the one hand, the speech generation models have given a voice to machines which are in turn helping humans with real-time speech translation~\cite{DBLP:journals/corr/abs-1910-00254}, removing language as a barrier in communication. 
The advances in Natural Language Processing (NLP) have enabled us to automate understanding a large amount of textual data.
On the other hand, computer vision and augmented reality have significantly impacted human life. They have provided better security surveillance and enriched gaming experience. Smart medical imaging systems are already assisting healthcare workers in providing a better diagnosis. AI has already indulged with human beings to a large extent which confirms that it is here to stay and grow. 

\section{What is Cognitive Science?}

Cognitive Science is the interdisciplinary study of the mind with a focus on computation. It is different from psychology, which focuses on the behavior, feelings, and thought processes of individuals and groups of people and addresses neuro-psych issues, mood disorders, personality, social psych, etc. It is also different from neuroscience which focuses on the neural apparatus that implements these computations. Despite the minor differences in focus, there is significant overlap in topics and approaches amongst cognitive science, psychology, and neuroscience. It is a highly interdisciplinary field involving psychology, linguistics, neuroscience, computer science, and philosophy. 

The roots date back far in history, but the real genesis of the interdisciplinary cognitive science field lies in the 1950s at the symposium of information theory (MIT). Alan Newell, Herbert Simon (computer scientists), Noam Chomsky (linguist), and George Miller (psychologist) presented work that each took a cognitive turn.

The word `Cognitive’ by itself means the mental process of knowing, understanding, and learning. Cognitive Science aims to understand the diverse approaches to understanding how the human mind works. It revolves around understanding the sources and principles of human intelligence, which in turn helps in the development and improvement of artificial intelligence.

\section{From AI to Cognitive Science}
The main focus of this paper is to bridge the gap between the two fields, which have independently proved useful for human civilization. We observe that understanding the brain is very important to develop better AI systems, and better AI systems are needed to have a clearer picture of how the brain works. We now discuss the various challenges faced by researchers, followed by some solutions to handle them.

\subsection{Challenges in building more human-like machines}
The task of developing a machine that can think and act like humans is humongous and daunting.
This is due to the ample search space of possible scenarios.
The current models requires extensive data even to learn a small task.
For example, object classification models are trained on millions of images for which they built statistical encoding.
In the case of a new image, they try to match it to the learned statistical encoding.
However, the human child does not need millions of images to recognize an object.
How the human brain is learning is still a black box.
This may sometimes lead to a question of whether some capabilities are encoded in our genes.
A human baby can recognize a face right from birth, which further boosts this theory.
Our capabilities of decoding genes are still minimal, and hence a lot of possible answers are still hidden.

Even after all the advances in neural networks and artificial intelligence, these systems still lack the flexibility and generality of the human mind.
AI systems are still not creative and do not perform well outside their domain.
Additionally, these systems do not have a sense of emotion, humor, and empathy.
Perhaps, the reason why we are not able to transmit these properties into the modern AI system is that we still do not understand how different humans feel these emotions.
We are still not able to get around the `Moravec's paradox'~\cite{moravec1988mind}. 

To add to this, there are ethical considerations involved in carrying out research on the brain. 
Researchers cannot directly intervene in the process of a healthy brain as inserting electrodes inside a healthy human brain is not permitted.
Most of the research is either being done on patients with epilepsy or animals.
Due to this there is bias in the data. Therefore, there is a strong need to find methods to study neural activity without penetrating the brain with electrodes.

Additionally, a significant challenge that is generally overlooked by the community is that both fields have grown enormously in complexity over the last few decades, and disciplinary boundaries have solidified. The large body of literature makes it even more challenging to work on multidisciplinary research without collaborating.

\subsection{A brain analogue}
The human brain is one of the most complex systems. A lot of pieces have to come together to explain how we are able to process and interpret information rapidly.
The computer model of the mind gives an analogy between the brain and computers. Computations and symbols best explain information processing in the brain and computers.
Information processing in the computer is done by programs operating on symbols, whereas information processing in the brain is the neural computations involving mental representations. In computers, we have symbols like binary numbers, whereas mental representations are symbols in the brain that have meaning or encode information.

The neurons in our brain inspire the artificial neural network. These networks consist of nodes that are analogous to neurons in brains. The connection between the nodes in the input layer and the hidden layers is made through mathematical weights, similar to synapses that connect two neurons in a brain. The synapses, in fact, are pretty complex in comparison to just numbers in a matrix. They use electrical and chemical activity to send a signal. Suppose an artificial neural network can produce a neural activity pattern that resembles a brain. In that case, we can study how neural network learns to understand the brain's functioning.

\subsection{Bridging the gap}
Researchers are continuously trying to use artificial intelligence for a better understanding of the brain. The artificial neural networks inspired by the networks of neurons that comprise the brain have completed an entire cycle. Now, cognitive science is taking advantage of AI to develop models to understand the brain and process complex data. The remarkable ability of neural networks to find deep hidden patterns in complex data is already helping researchers. Functional magnetic resonance imaging (fMRI) produces very complex neural brain activity data at a resolution of 1–2 millimeters every second. Cohen, Jonathan D., et al.~\cite{cohen2017computational} have effectively used neural networks to find signals in the image that are very, very large.  Brain-Machine Interface is being used to map computers from the human brain. One such work by Pandarinath et al.~\cite{pandarinath2021science} uses AI to understand brain signals. They gather the brain-activity recordings of just 200 neurons for arm movement and fed them to an artificial neural network. The model successfully finds the latent factors or patterns in that data. This helped the researchers in predicting the arm movement of an animal on a millisecond-by-millisecond basis. Such data is precisely what is needed to control a robotic arm.

Afraz, A., Yamins, D. L., \& DiCarlo, J. J.~\cite{afraz2014neural} proposed a system that could reproduce brain data. They created a deep neural network based on the ventral visual stream and trained it with thousand of images of $64$ objects. They showed similar objects to a monkey and recorded the neural activity. They found out that the best neural networks produce the same neural activity patterns as generated by the brain, with a $70\%$ similarity. Yamins~\cite{kell2018task} extended this research to the auditory cortex by training a deep neural network from $2$-second clips. It helped the researchers identify the areas of the brain which perform speech recognition and which perform music recognition. These studies show that if the researchers can develop a model that performs similarly to the brain, it will give us an idea of how the brain is doing the same task. This can provide at least one possible hypothesis of how the brain functions.

More recent research work to better understand the human brain using AI includes reverse engineering of the infant's brain to know how language is learned in the early phase of development (Dupoux, E~\cite{dupoux2018cognitive}).

Cognitive science is also helping in the improvement of neural networks. CognitiveCNN was proposed to reconstruct, and classify the images, which was inspired by feature integration theory in psychology to utilize human-interpretable features like shape, texture, edges, etc.(Mohla, Satyam, et al.~\cite{mohla2020cognitivecnn}). The idea was further extended to Convolutional Neural Networks (CNNs) for achieving superior performance in object recognition. Russin et al.~\cite{russin2020deep} argued that the current challenges faced by the state-of-the-art artificial intelligence systems could be tackled by using the computational principles thought to be at work in the prefrontal cortex. The inductive biases and principles at work in the prefrontal cortex may inspire novel architectures. Active inference, an emerging framework within cognitive and computational neuroscience, has augmented our approach for traditional Reinforcement Learning by balancing exploration and exploitation (Tschantz, Alexander, et al.~\cite{tschantz2020reinforcement}).

It is well known in the scientific community that neural networks struggle to understand even the basic physical intuitions. Piloto, Luis S., et al.~\cite{pilotolearning} has used notions of core object knowledge from developmental psychology to predict the future of $3$D physical scenes from segmented images. Lingelbach, Michael John, et al.~\cite{lingelbach2020towards} built a self-learning system based on human cognitive approaches to learn physical dynamics.

Moving further in improving artificial neural networks using human cognitive behavior, an interesting paper by Tadros T et al.~\cite{tadrossimulated} exploits the hypothesis that sleep promotes generalization of learned examples in mammals. They showed that spiking neural network trained with spike-timing-dependent plasticity gives better generalization and robustness on novel inputs when an offline, sleep-like period is used after training.

Cognitive Science and AI would fit together as they are basically focused on the same thing. On the one hand, the target is to find a mathematical model to solve a learning problem so that it can be used for a machine. On the other hand, the target is to study the only known solution to that problem, i.e., the human brain. There are a couple of advantages in developing AI around cognitive science. Firstly, neuroscience provides a lot of new inspiration for architectures and algorithms. These are not even bounded by the mathematical and logic-based methods, which are majorly used in current AI systems. Secondly, if a known algorithm is found working in the brain, then it acts as a validation for the current AI model.

\section{Will AI beat human intelligence ever?}
What if AI is able to learn things by itself and make decisions? What if AI starts killing humans by considering them as a threat? These types of questions have crossed the mind of many of us. Some of the pioneers of the 21st century, like Elon Musk and Stephen Hawkins, have also pointed out the potential danger of the uncontrolled growth of AI systems. Once the computers were able to program themselves and achieve the `technology singularity', the risk of a potential fight between machines and humans for resources can not be ignored. 

The current progress in AI has even fueled this fear. The rise of `Deep Fakes' and their ability to generate realistic human-like faces is one such example~\cite{nguyen2019deep}. They pose a real threat to online privacy and can easily be used to blackmail someone. They are already fooling facial recognition systems~\cite{korshunov2018deepfakes}. At the same time, research is being made to understand how they work and hence finding a way to beat them~\cite{jung2020deepvision}. 
These systems are still limited to the human head and terribly fail to generate a complete human body. 
There are some positive applications as well like they are beneficial in creating low-bandwidth video calls~\cite{wang2021one}. 

The majority of the AI systems used today have some form of bias encoded in them.
For example, most facial recognition systems used for critical tasks like searching for a terrorist are trained on the faces of white people and single out black people as the threat.
This goes to show that AI systems have a long way in becoming as intelligent as humans in assessing the situation and reacting to it.
Additionally, the companies creating and deploying such systems need to be extra careful with what kind of data they are feeding into these algorithms and the system's different biases.
Knowing the limitations and keeping them under control will help create responsible AI that won't hurt humanity once it is fully developed.

Since the industrial revolution, the automation of mundane tasks has led to the rapid advancement of society. To add to it, recent advances in computer sciences, particularly in artificial intelligence, have led to a different form of revolution that will further advance society.
However, unlike the industrial revolution, which only had a negative impact on the environment, this revolution will have environmental and psychological impacts.
As people become more dependent on digital automation, mindless scrolling, and window shopping, they tend to use their brains less, which leads to diseases like Alzheimer's in the latter part of their lives.
Additionally, many businesses have been impacted by the automation of the processes. The people of those businesses are not equipped enough to learn about the new technology and use it for their benefit.
On the other hand, training and deploying such a system requires a massive amount of computation resources, which leads to a large carbon footprint.
Due to this, allowing the AI systems to overtake the world for us could be a tricky thing to do.
We should be more mindful in deciding which part of the processes we are automating and how it will impact in the long run.

The limitations of AI to understand emotions are also well known to the community. 
Even after all these advances, emotions in the AI system have not been achieved. This puts a significant doubt on the ability of AI to come even closer to human cognitive abilities. The areas like graphic designing, human resource management in organizations, music, and other artistic industries, etc., still need the human touch. Unless an AI system can integrate emotional intelligence, self-awareness, and experience, the possibility of an AI takeover is minuscule.

Some researchers question `smart' AI’s ability to wage wars or be useful for human conflicts. This would be the antagonized version of an AI that has learned war games by itself.
Due to this, the usage of AI systems for warfare should be highly regularised.
We do not want algorithms, which do not have emotions, to decide to drop bombs at any part of the world.
We need to be more responsible and respectful to human intelligence for such crucial tasks.

\section{`Smart' AI: The possibilities}
As the research advances and the field of cognitive science comes closer to artificial intelligence, it will redefine how humans interact with machines.
Every industry has to understand what will work for the human mind to get the benefit. Understanding the mental and emotional processes in human beings is a must for any system that directly interacts with a human. 

For example, the education sector could utilize AI to comprehend the necessities of students and embrace innovation to train them better. Using animations automatically created by generative AI systems could benefit in teaching subjects better at significantly fewer costs.
Creating systems that analyze the understanding of the subject being taught in real-time could help better teach the students and create a better workforce.
Such systems could also help identify weak students and provide them additional attention to help them perform better in life.

Furthermore, smart AI models can revolutionize the medical sector. It helps in diagnosing diseases at a very early stage and can provide a better cure. It will help disentangle the brain and can provide control over our ability to learn and store information, i.e., control over memory. The better mapping of brain signals will help deliver bionic limbs to the people who lost them due to some misfortune. In many cases, the inability to understand brain signals using current technologies has prevented doctors from helping humans born with a disability. A recent work~\cite{pandarinath2019brain} by researchers at Emory University, and Georgia Institute of Technology has shown a possibility of developing a device that can generate voice using brain signals. It could be a stepping stone to restoring speech function in individuals unable to speak.

`Smart' AI can also open the doors to explore the uncharted territories of outer space. Humans cannot travel for a large duration in zero gravity because of physical limitations. This can be solved by sending smart robots to explore these areas. The decision-making capabilities of such systems will help in tackling any unforeseen challenges that are pretty common in outer space. 

\section{Conclusion}
AI is bound to integrate into the human world with a lot of effort, first conquering the ability to think independently and with empathy. While scientists will use cognitive science to replicate intelligence in AI, AI will eventually be using cognitive science to understand humans better. It will develop the ability to learn new things, make decisions by rationalizing, and provide valuable insights for human beings. AI should be viewed as a constructive tool toward understanding what it means to be human. 
It can be considered as a  constructivist paradigm in humanity’s long-standing quest for self-discovery.
The fear of artificial general intelligence will be significantly reduced if we better understand how they work. The key to it lies in a better understanding of the human brain. The future is `smart' for AI and easier for us. We must never forget: ``With great power comes great responsibility".

\bibliographystyle{plain}
\bibliography{bibliography}

\end{document}